\documentclass[10pt,twocolumn,letterpaper]{article}

\usepackage[pagenumbers]{cvpr} 
\usepackage{float}
\usepackage[ruled,vlined]{algorithm2e}
\usepackage{algorithmic}
\usepackage{multirow}
\usepackage{booktabs}
\usepackage{bbding}

\usepackage[dvipsnames]{xcolor}

\definecolor{cvprblue}{rgb}{0.21,0.49,0.74}
\usepackage[pagebackref,breaklinks,colorlinks,citecolor=cvprblue]{hyperref}

\title{Continual Self-supervised Learning: Towards Universal Multi-modal Medical Data Representation Learning}

\author{
Yiwen Ye$^{1}$~~~Yutong Xie$^{2}$\Envelope~~~Jianpeng Zhang$^{1}$~~~Ziyang Chen$^{1}$~~~Qi Wu$^{2}$~~~Yong Xia$^{1}$\Envelope\\
$^{1}$ School of Computer Science and Engineering, Northwestern Polytechnical University, China\\
$^{2}$ The University of Adelaide, Australia\\
{\tt\small ywye@mail.nwpu.edu.cn, yutong.xie678@gmail.com, jianpeng.zhang0@gmail.com} \\ 
{\tt\small zychen@mail.nwpu.edu.cn, qi.wu01@adelaide.edu.au, yxia@nwpu.edu.cn}
}

\begin{document}
\maketitle
\begin{abstract}
Self-supervised learning is an efficient pre-training method for medical image analysis. However, current research is mostly confined to specific-modality data pre-training, consuming considerable time and resources without achieving universality across different modalities. 
A straightforward solution is combining all modality data for joint self-supervised pre-training, which poses practical challenges. 
Firstly, our experiments reveal conflicts in representation learning as the number of modalities increases. 
Secondly, multi-modal data collected in advance cannot cover all real-world scenarios.
In this paper, we reconsider versatile self-supervised learning from the perspective of continual learning and propose MedCoSS, a continuous self-supervised learning approach for multi-modal medical data. Unlike joint self-supervised learning, MedCoSS assigns different modality data to different training stages, forming a multi-stage pre-training process. 
To balance modal conflicts and prevent catastrophic forgetting, we propose a rehearsal-based continual learning method. 
We introduce the $k$-means sampling strategy to retain data from previous modalities and rehearse it when learning new modalities. 
%
Instead of executing the pretext task on buffer data, a feature distillation strategy and an intra-modal mixup strategy are applied to these data for knowledge retention.
We conduct continuous self-supervised pre-training on a large-scale multi-modal unlabeled dataset, including clinical reports, X-rays, CT scans, MRI scans, and pathological images. Experimental results demonstrate MedCoSS's exceptional generalization ability across nine downstream datasets and its significant scalability in integrating new modality data.
Code and pre-trained weight are available at \href{https://github.com/yeerwen/MedCoSS}{https://github.com/yeerwen/MedCoSS}.
\end{abstract}

\renewcommand{\thefootnote}{}
\footnotetext{
Corresponding authors: Yong Xia and Yutong Xie. 
}

\begin{figure}[t]
  \centering
   \includegraphics[width=0.99\linewidth]{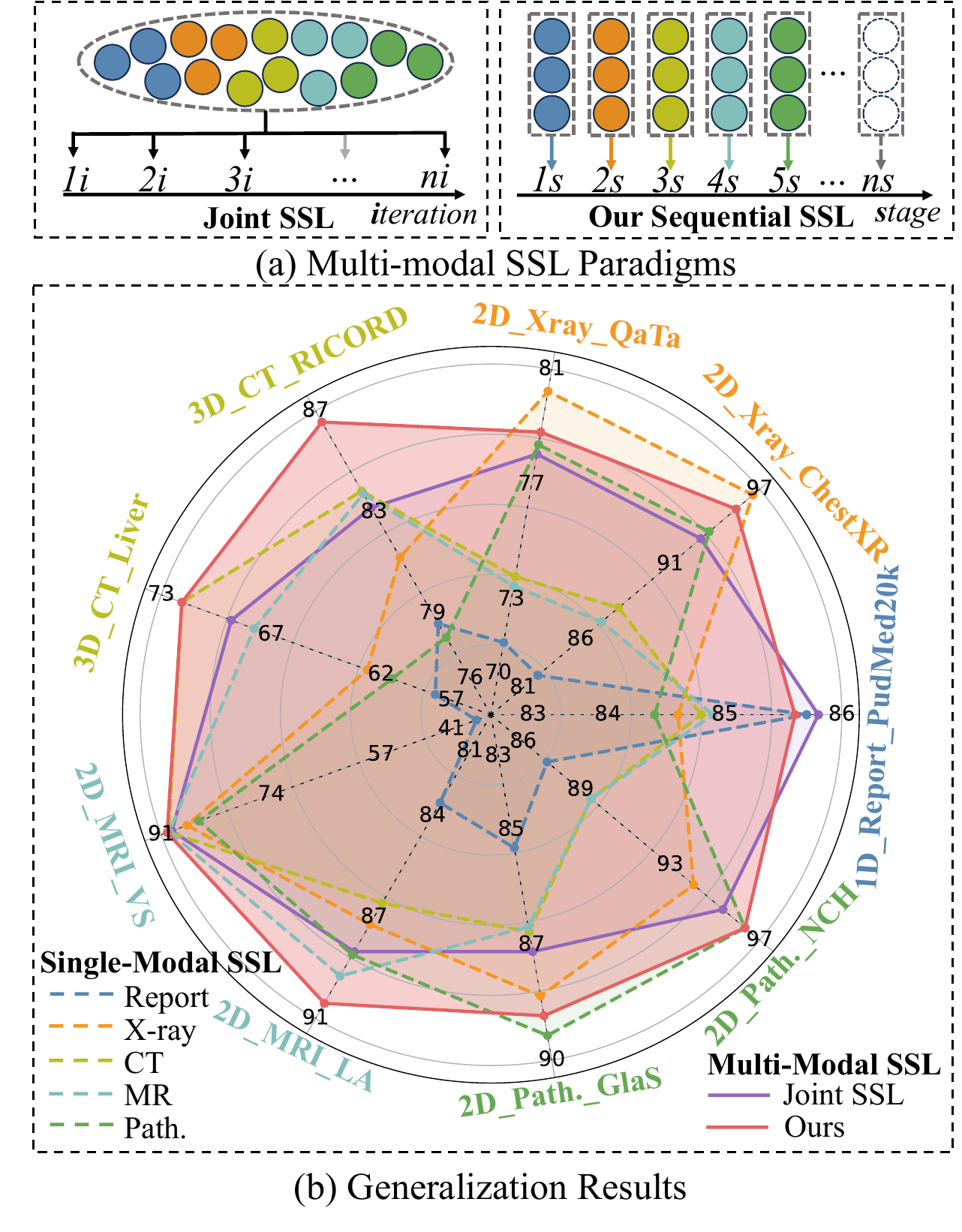}
   \caption{
   (a) Comparison between joint SSL and our sequential SSL. Different colored balls represent different modalities. The joint SSL suffers from the high costs to integrate knowledge from new data while our sequential SSL shows a distinct advantage in this aspect.
   (b) Generalization performance of five single-modal pre-trained models (dashed lines) and two multi-modal pre-trained models (solid lines). 
   The label shows the data dimension, modality, and dataset name. 
   The model from MedCoSS achieves state-of-the-art generalization performance (red solid line) on a broad range of tasks, beating other models.
   }
   \label{fig: intro}
\end{figure}
\section{Introduction}
\label{sec:intro}


Medical self-supervised learning (SSL) emerges as a powerful alternative to large-scale supervised learning, showcasing impressive feats in high-quality representation learning. Its essence lies in pre-training strong encoder or encoder-decoder architectures for subsequent downstream tasks by solving a pretext task without human annotations.
A review of existing medical SSL approaches highlights a pattern: they primarily focus on a single modality data, such as X-rays \cite{bozorgtabar2020salad,  azizi2021big, zhou2021preservational, zhou2021models, haghighi2022dira, taher2022caid, cai2023dual}, computed tomography (CT) scans \cite{zhu2020rubik, zhou2021preservational, zhou2021models, haghighi2022dira, ye2022desd, tang2022self, yan2023localized, zhang2023dive, jiang2023anatomical}, magnetic resonance image (MRI) scans \cite{cai2023dual, ouyang2022self, zhou2022dual, kalapos2022self}, pathological images \cite{kang2023benchmarking, wang2021transpath, yang2021self}, fundus images \cite{li2021rotation}, multivariate cardiac signals \cite{lan2022intra}, and ultrasound (US) images \cite{chen2022generating}, or only a handful of paired modality data like radiology reports paired with X-rays \cite{chen2022multi,yan2022clinical,liu2023improving,liu2023m,zhang2023knowledge}, MRI modality pairs \cite{taleb2021multimodal}, and images paired with genomics \cite{jin2023gene}.
This narrow focus constrains them from extending to the universal medical pre-training since a large-scale paired multi-modal dataset is often impractical.
Despite a few pioneering attempts in unpaired multi-modal pre-training \cite{xie2022unimiss, cai2022uni4eye, ghesu2022self}, the scope of these efforts remains notably constrained. Specifically, these efforts are either limited to only two modalities \cite{xie2022unimiss}, only for a specific field \cite{cai2022uni4eye}, or restricted to a specific dimension \cite{ghesu2022self}. 
These limitations hinder their ability to delve deep into the challenges of universal multi-modal SSL in the face of more modalities.

In our experiments, we used the plain ViT/B \cite{dosovitskiy2020vit} as the backbone and masked modeling \cite{he2022masked, devlin2018bert} as the pretext task. The results are the average values of accuracy, AUC, and F1 for classification tasks or Dice for segmentation tasks. We observed a consistent trend: single-modal pre-trained models excel in downstream tasks with identical modality but falter considerably when dealing with the data from another modal (see Fig. \ref{fig: intro}(b)).
This observation underscores the urgency to craft a universal pre-training model proficiently handling varied medical multi-modal data.
A seemingly intuitive solution to this problem is joint SSL (see Fig. \ref{fig: intro}(a)), where all accessible medical data (from various modalities) are completely collected and involved in pre-training executed as single-modal per mini-batch considering the diverse dimensions inherent in medical imaging \cite{xie2022unimiss, cai2022uni4eye}.
However, compared to single-modal SSL, where the models are pre-trained and fine-tuned on an identical modality, joint SSL falls short of expectations, despite its richer information base. This is evident from Fig. \ref{fig: intro}(b), which shows a significant performance drop in most downstream tasks.
We attribute this performance degradation to the phenomenon, which we term as \textit{modal data collision}, that the representation learning on two modalities conflicts with each other owing to the vast disparities between those modalities.
Moreover, a joint training paradigm is also inadvisable, due to the high costs associated with incorporating the knowledge from both the seen and newly introduced modal data, necessitating a full reiteration of the pre-training process.

To circumvent the obstacle of \textit{modal data collision} and recover the scalability of the pre-trained model, we propose the \textbf{Med}ical \textbf{Co}ntinual \textbf{S}elf-\textbf{S}upervised (MedCoSS) paradigm for multi-modal representation learning. MedCoSS diverges from traditional joint pre-training by adopting a sequential pre-training paradigm \cite{riemer2018learning}, assigning each stage to a specific modality. 
To avoid catastrophic forgetting, we design a rehearsal-based continual learning technique that retains a subset (\textit{e.g.}, 5\%) of previous modal data within a rehearsal buffer, using the $k$-means sampling strategy over random sampling. Specific to previous knowledge retention, we deploy a feature distillation strategy and intra-modal mixup strategy during subsequent pre-training stages.
According to our MedCoSS paradigm, a model undergoes pre-training on the data from five modalities: Report, X-ray, CT, MRI, and Pathological imaging. Evaluation on nine downstream datasets encompassing all seen modalities at the pre-training phase to showcase the superior generalization capability of our model. 
The model from MedCoSS consistently outperforms the models developed through single-modal pre-training, joint pre-training, and other multi-modal pre-training paradigms, while maintaining cost-effective scalability to include new knowledge, paving the way to build the multi-modal pre-trained medical universal model.
%
In summary, our contributions can be encapsulated as:
\begin{itemize}
    \item We identify and mitigate the \textit{modal data collision} issue and innovate the MedCoSS paradigm. By shifting from joint to sequential training and introducing continual learning, we alleviate the collisions and cost-effectively scale new knowledge without forgetting the old.
    \item We conduct an in-depth exploration into unpaired multi-modal SSL, expanding the modality and data dimension. We integrate five prevalent modalities, including the Report, X-ray, CT, MRI, and Pathological imaging, spanning three dimensions (1D, 2D, and 3D) to pre-train a universal model using the proposed MedCoSS.
    \item The model developed through our MedCoSS achieves state-of-the-art generalization performance on a broad range of downstream tasks, indicating a potential direction to develop the multi-modal pre-trained medical universal model.

\end{itemize}

\begin{figure*}[t]
  \centering
   \includegraphics[width=1.0\linewidth]{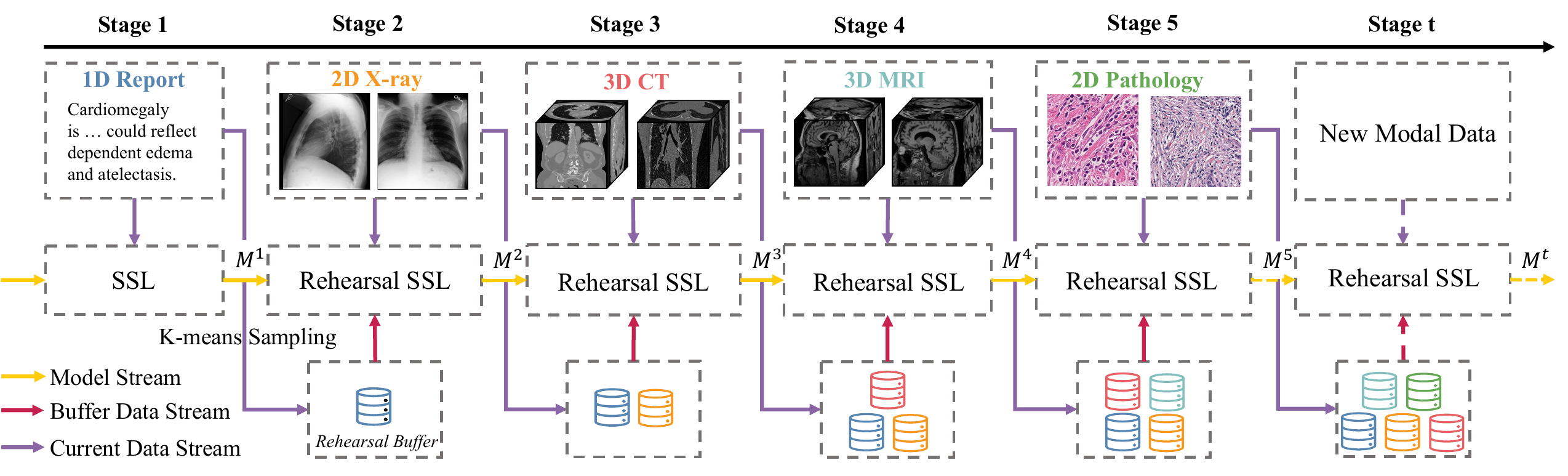}
   \caption{Pipeline of proposed MedCoSS paradigm. MedCoSS leverages a sequential pre-training manner to produce a multi-modal pre-trained model ($M^{t}$) through multi-stage learning, each assigned to specific modal data. To alleviate the issue of catastrophic forgetting, a rehearsal buffer is adopted and maintained for previous modal data rehearsal in the subsequent pre-training. After each stage, the model $M_{t}$, which consists of three tokenizers and a Transformer-based encoder, will be used to select samples from current data using a $k$-means sampling strategy and then transferred to the next stage.
  }
   \label{fig: overview}
\end{figure*}

\section{Related Work}
\label{sec:related work}

\subsection{Medical self-supervised learning}
Medical SSL has been increasingly used to improve the performance of medical image analysis, achieving success in various clinical applications, with its success largely dependent on large-scale pre-training data.
The modalities of data directly influence the performance of models in target tasks. 
Generally, popular medical SSL can be categorized into three primary categories: single-modal SSL, paired multi-modal SSL, and unpaired multi-modal SSL.
\textbf{Single-modal SSL}: this approach leverages only one modality for pre-training. Typical modalities in this category encompass X-ray \cite{bozorgtabar2020salad,  azizi2021big, zhou2021preservational, zhou2021models, haghighi2022dira, taher2022caid, cai2023dual}, CT \cite{zhu2020rubik, zhou2021preservational, zhou2021models, haghighi2022dira, ye2022desd, tang2022self, yan2023localized, zhang2023dive, jiang2023anatomical}, MRI \cite{cai2023dual, ouyang2022self, zhou2022dual, kalapos2022self}, Pathological imaging \cite{kang2023benchmarking, wang2021transpath, yang2021self}, fundus image \cite{li2021rotation}, multivariate cardiac signals \cite{lan2022intra}, and ultrasound \cite{chen2022generating}. Despite the versatility of single-modal SSL, our observations in Fig. \ref{fig: intro}(b) indicate that models trained on one modality often struggle to generalize across other modalities, particularly when these modalities differ in dimensionality.
\textbf{Paired multi-modal SSL}: this approach leverages multiple modalities with an inherent relationship for pre-training. The design purpose behind the corresponding pretext tasks is to encourage the model to be aware of these relationships. Existing SSL paradigms employ various data pair types, such as clinical reports paired with images \cite{chen2022multi,yan2022clinical,liu2023improving,liu2023m,zhang2023knowledge,wu2023towards}, captions paired with images \cite{wu2023towards}, or MRI modality pairs \cite{taleb2021multimodal}. However, collecting such paired multi-modal datasets is resource-intensive and impractical, even for widely used modalities.
\textbf{Unpaired multi-modal SSL}: this approach presents a more economical alternative to paired multi-modal SSL by relaxing data constraints, allowing for easier scaling.
Pioneering efforts have delved into this field.
Ghesu \textit{et al.} \cite{ghesu2022self} collected a 2D mixed modality dataset comprising X-rays, CT slices, MRI slices, and US images. Their approach, focusing on multi-modal clustering, enabled distinct online clustering for each modality using individual prototypes.
Cai \textit{et al.} \cite{cai2022uni4eye} devoted to ophthalmology, creating a multi-modal dataset and leveraging masked image modeling (MIM) for effective representation learning.
Xie \textit{et al.} \cite{xie2022unimiss} introduced an iterative training scheme for joint training of CT and X-ray data, incorporating volume-slice consistency as a form of cross-dimension regularization during CT pre-training.
These studies, each addressing different aspects of unpaired multi-modal SSL, collectively enrich our understanding of this field.
However, as modalities and data dimensions increase, these paradigms often fall short in fine-tuning compared to single-modal pre-training, hindered by the modal data collision in joint training.
Crucially, the pre-trained models obtained from these paradigms exhibit limited scalability when it comes to incorporating knowledge from new data sources.
Therefore, we introduce continual learning technology in the SSL paradigm to mitigate these issues, proposing a more practical and efficient approach to multi-modal SSL.

\subsection{Continual learning for knowledge retention}
Continual Learning (CL) techniques have been instrumental in aiding models to retain knowledge over time, especially when sequentially exposed to diverse data or tasks. 
These CL techniques can be summarized in five main categories, including regularization-based approach \cite{kirkpatrick2017overcoming}, rehearsal-based approach \cite{buzzega2020dark,riemer2018learning}, optimization-based approach \cite{lin2022towards}, representation-based approach \cite{fini2022self}, and architecture-based approach \cite{mallya2018packnet}.
The rehearsal-based approach, notably successful in retaining knowledge when equipped with a rehearsal buffer, usually maintains some previous data \cite{riemer2018learning} or features \cite{buzzega2020dark} and replays them during subsequent learning. 
A standout method within this category is DER \cite{buzzega2020dark}, which retains previous training samples and their associated logits, focusing on logit-matching rather than label-matching.
In this study, we adjust DER with three customized modifications for our MedCoSS, including:
(1) we replaced reservoir sampling with a $k$-means sampling strategy, considering sample diversity; 
(2) instead of directly preserving logits of samples, we only preserved samples and utilized a freeze encoder to extract logits, catering to the random masking operation for inputs;
(3) we introduced an intra-modal mixup strategy to augment data from the rehearsal buffer.


\begin{figure}[t]
  \centering
   \includegraphics[width=1.0\linewidth]{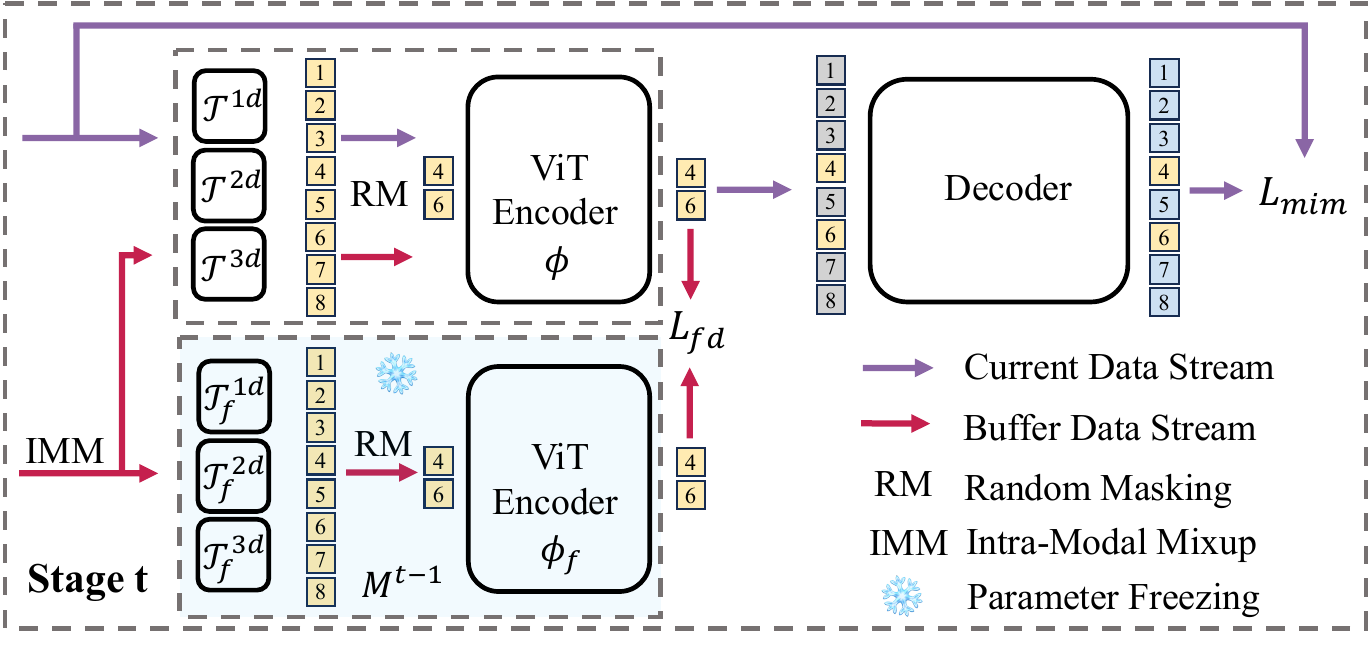}
   \caption{
   Illustration of proposed rehearsal SSL paradigm. The masked modeling pretext task \cite{he2022masked, devlin2018bert} is used for representation learning on current modal data corresponding to stage $t$. The feature distillation strategy is used for knowledge retention by building a freeze model $M^{t-1}$ and rehearsal buffer. 
   }
   \label{fig: ssl}
\end{figure}
\section{Approach}
\subsection{Overview}
The proposed MedCoSS paradigm is designed as a two-step SSL, which comprises an unsupervised pre-training phase and a fully-supervised fine-tuning phase.
During pre-training, we employ masked image/language modeling as the pretext task to extract generalized representations from an integrated set of multi-modal data, specifically clinical reports, X-rays, CT scans, MRI scans, and pathological images. 
To circumvent the obstacle of \textit{modal data collision} caused by the joint multi-modal pre-training, we introduce a sequential pre-training scheme, in which each stage is tailored for training on a specific modality of data.
The potential risk of knowledge forgetting during this sequential process is countered with continual learning techniques.
During fine-tuning, the pre-trained encoder is paired with a randomly initialized task-specific head for each downstream task.
A comprehensive visualization of our MedCoSS paradigm was illustrated in Fig. \ref{fig: overview}. We now delve into the details of each part.

\subsection{Universal architecture for multi-modal SSL}
We craft a universal architecture \cite{zhu2022uni} in line with the aim of universal multi-modal SSL.
The medical data from various modalities could be either 1D (\textit{e.g.}, clinical reports), 2D (\textit{e.g.}, X-rays and pathological images), or 3D (\textit{e.g.}, CT and MRI scans).
To accommodate these 
data seamlessly, a dimension-free backbone is imperative.
For this study, we employ three dimension-specific tokenizers to convert 1D, 2D, and 3D medical data into token sequences, respectively, and use a plain ViT/B \cite{dosovitskiy2020vit} as the encoder for representation learning in a sequence-to-sequence manner regardless of the dimension of the medical data.
Specifically, we deploy the byte pair encoding (BPE) tokenizer \cite{sennrich2015neural} for text and 2D/3D image patch tokenizer \cite{dosovitskiy2020vit} for 2D/3D visual data.
For text, following BERT \cite{devlin2018bert}, we mask out 15\% words randomly. The model predicts each masked word according to the visible words, using the cross-entropy loss as a constraint.
For visual data, following MAE \cite{he2022masked}, token sequences 
are randomly masked at a ratio of 75\% and only those unmasked token sequences are fed to the encoder. Subsequently, the encoded visible token sequences, together with learnable mask tokens, serve as the inputs of the Transformer-based decoder, aiming to reconstruct the previously masked tokens. The mean squared error (MSE) loss is employed to ensure a high consistency between the original and reconstructed images in masked regions.

\subsection{Rehearsal-based continual pre-training}
The typical approach to joint pre-training on multi-modal data often encounters the issues of \textit{modal data collision} and high cost of integrating new knowledge.
To address both issues, we advocate a shift towards a sequential pre-training paradigm, effectively assigning each stage to a specific imaging modality.
Formally, considering $T$ unlabeled subsets of data $D=\{D_{1}, D_{2},..., D_{T}\}$, each being acquired from a unique modality.
The correspondence between modalities and stages is random.
Unlike the standard practice of directly pre-training a model $M$ on $D$, our MedCoSS paradigm pre-trains $M$ sequentially on each subset $D_{t}$ in the $t^{th}$ stage, where the intermediate pre-trained model is denoted by $M^{t}$.
This paradigm strategically circumvents \textit{modal data collision} by isolating different modalities during pre-training while at the risk of catastrophic forgetting.
We counter this risk by integrating rehearsal-based continual learning techniques to preserve previous knowledge.
When stage $t$ unfolds, the pre-training not only focuses on the MIM pretext task using $D_{t}$, but also engages an auxiliary feature distillation task to prevent catastrophic forgetting. 

\noindent{\textbf{Learn from current modality}. We utilize $M_{t-1}$, which includes an encoder $\phi$ and three tokenizers $\mathcal{T} ^{1d}, \mathcal{T}^{2d}, \mathcal{T}^{3d}$, along with a randomly initialized decoder to continually learn new knowledge from current modal data on a masked modeling pretext task. 

\noindent{\textbf{Learn from previous modalities}. We establish a rehearsal buffer $B=\{x_{i}^{j}\mid x_{i}^{j} \in  D_{j}, j \in (1, 2,..., t-1) \}$ to retain a fraction of training data from each of previous stages (see Section \ref{method: kmeans} for details). 
%
%
Moreover, we copy a frozen version of $M_{t-1}$, where an encoder and three tokenizers are denoted as $\phi_{f}, \mathcal{T}_{f}^{1d}, \mathcal{T}_{f}^{2d}, \mathcal{T}_{f}^{3d}$, respectively.
For each sample $x$ from $B$, the intra-modal mixup (IMM) strategy is utilized (see Section \ref{method: IMM} for details) for augmentation. 
The augmented sample is processed by the learnable network (\textit{i.e.}, $\mathcal{T}^{1d}, \mathcal{T}^{2d}, \mathcal{T}^{3d}$, and $\phi$) and frozen network (\textit{i.e.}, $\mathcal{T}_{f}^{1d}, \mathcal{T}_{f}^{2d}, \mathcal{T}_{f}^{3d}$, and $\phi_{f}$), respectively. The embeddings produced by both networks are then encouraged to be similar via minimizing the MSE loss (see $L_{fd}$ in Fig. \ref{fig: ssl}).


The pre-training process of our MedCoSS paradigm is summarized in Algorithm \ref{alg: main}.

\SetKwInOut{Require}{Require}
\SetKwInOut{Initialize}{Initialize}
\begin{algorithm}[t]
\KwIn{$T$ datasets $\{D_{1},...,D_{T}\}$ with different modalities, rehearsal buffer $B$, tokenizers $\mathcal{T}$, encoder $\phi$, model-specific decoders $\psi_{i},(i \in (1,...,T))$, sampling operation $Sample(.)$}
\KwOut{$\phi$ and $\mathcal{T}$}
\caption{MedCoSS's pre-training algorithm.}
\begin{algorithmic}[1]
\STATE $\triangleright$ \textbf{Stage 1}:
    \STATE Training dataset $D \leftarrow D_{1}$ 
    \STATE Update $\phi$, $\mathcal{T}$, and $\psi_{1}$ by $L_{mim}$
    \STATE $B \leftarrow Sample(D_{1})$
\STATE $\triangleright$ \textbf{Stage 2, 3, ...}:
\FOR {\textit{i} = 2, 3, ...}
    \STATE Training dataset $D \leftarrow D_{i} \cup B$ 
    \FOR {\textit{iteration} = 1, 2, ...}
        \STATE Sample a batch of unlabeled data $x$ from $D$
        \IF{$x \in D_{i}$} 
            \STATE Update $\phi$, $\mathcal{T}$, and $\psi_{i}$ by $L_{mim}$
        \ELSE
            \STATE Update $\phi$ and $\mathcal{T}$ by $L_{fd}$
        \ENDIF
    \ENDFOR
    \STATE $B \leftarrow B \cup Sample(D_{i})$
\ENDFOR 
\end{algorithmic}
\label{alg: main}
\end{algorithm}

\subsubsection{Rehearsal buffer construction} \label{method: kmeans}
In our MedCoSS paradigm, the rehearsal buffer $B$ is constructed for subsequent training stages by using the $k$-means sampling strategy
to select representative samples from each subset $D_t$. Unlike the random sampling that lacks reliability in capturing data diversity, $k$-means sampling consists of two steps: (1) clustering each subset $D_t$ into $C$ clusters based on the embeddings produced by the pre-trained model (\textit{i.e.}, $M_{t-1}$ in the $t^{th}$ Stage), and (2) selecting $K$ samples, which are nearest to the center, from each cluster. Note that the number of clusters $C$ is empirically set to 1\% of the subset size.

\subsubsection{IMM Augmentation Strategy} \label{method: IMM}
In MedCoSS, the IMM augmentation strategy enhances the variety of the samples drawn from the rehearsal buffer $B$ to deal with a limited buffer size for each modality.

\noindent{\textbf{For text data}},
we denote the batch of text samples as $b \in \mathbb{R}^{N \times L}$, where $N$ represents the batch size and $L$ means the length of text sequences.
To perform augmentation, we initiate by copying the batch $b$ to produce $b^{'}$, which is subsequently shuffled to create a new sequence order. A binary mixup strategy is then employed to synthesize a new batch $b^{mix}$, defined as:
\begin{equation}
b^{mix} = \lambda_{b} \odot b + (1 - \lambda_{b}) \odot b^{'}, 
\label{eq1}
\end{equation}
where $\odot$ denotes element-wise multiplication, and $\lambda_{b}$ represents a binary mask array matching the shape of $b$. 
This mask's values, either 0 or 1, are derived from an array generated by uniformly sampling numbers with a random threshold, also sourced from a uniform distribution within the range of $[0, 1)$.

\noindent{\textbf{For visual data}},
the augmentation process begins similarly by duplicating and shuffling the image batch $b$ to obtain $b^{'}$.
The difference is that we incorporate a continual mixup strategy here, expressed as:
\begin{equation}
b^{mix} = \lambda_{c} \odot b + (1 - \lambda_{c}) \odot b^{'}, 
\label{eq2}
\end{equation}
where the continual mask vector $\lambda_{c}$ is an array of length $N$ with values randomly chosen between $[0, 1)$. Each number in this vector corresponds to a specific sample, determining how much each image in batch $b$ blends with its counterpart in batch $b^{'}$.

\subsection{Fine-tuning on downstream tasks}
After pre-training, we adopt the pre-trained encoder and customize it for various downstream tasks, including the classification and segmentation of all seen modalities. 
Given a task, a corresponding prediction head is designed according to the dimensionality of input data and the task type.
We use the Multi-Layer Perceptron (MLP) head for classification tasks and employ a convolutional-based decoder with a segmentation head for segmentation tasks. More details are in Appendix \ref{sec:model details}.


\begin{table}[t]
  \centering
  \caption{Overview of the datasets used in this study.}
  \resizebox{1\columnwidth}{!}{
    \begin{tabular}{c|c|c|c|ccc}
        \toprule
          & Dataset & Modality  & Task  & \#Train & \#Val & \#Test \\
      \midrule
    \multirow{5}[0]{*}{SSL} & MIMIC-CXR & Report & \multirow{5}[0]{*}{MLM/MIM} & 227,323 & -     & - \\
          & MIMIC-CXR & X-ray &       & 356,309 & -     & - \\
          & DeepLesion & CT    &       & 10,594 & -     & - \\
          & ADNI  & MRI   &       & 3,789 & -     & - \\
          & TCGA  & Path. &       & 217452 & -     & - \\
        \midrule
    \multirow{9}[0]{*}{DS} & PudMed20k & Report & Cls   & 180,040 & 30,212 & 30,135 \\
          & ChestXR & X-ray & Cls   & 14,365 & 3,593 & 3,432 \\
          & QaTa  & X-ray & Seg   & 5,716 & 1,429 & 2,113 \\
          & RICORD & CT    & Cls   & 194   & 52    & 84 \\
          & LiTS  & CT    & Seg   & 104   & -     & 27 \\
          & VS    & MRI   & Seg   & 193   & -     & 49 \\
          & LA    & MRI   & Seg   & 80    & -     & 20 \\
          & NCH   & Path. & Cls   & 79,994 & 20,006 & 7,180 \\
          & GlaS  & Path. & Seg   & 67    & 18    & 80 \\
        \bottomrule
    \end{tabular}%
    }
  \label{tab:data}%
\end{table}%

\section{Experiment}
\subsection{Datasets}
The datasets used in this study can be divided into upstream unlabeled and downstream labeled datasets. The summary of these datasets is shown in Table \ref{tab:data}.

\noindent{\textbf{Upstream datasets}}:
Our large-scale multi-modal unlabeled dataset encompasses five modalities: Report, X-ray, CT, MRI, and Pathological imaging (Path.). The clinical reports and X-rays are sourced from the JPG version of the MIMIC-CXR 2.0.0 dataset \cite{johnson2019mimic}.
CT scans are obtained from the DeepLesion dataset \cite{yan2018deeplesion}.
MRI scans are drawn from the Alzheimer’s Disease Neuroimaging Initiative (ADNI) database, including ADNI-1, ADNI-2, and ADNI-GO datasets \cite{jack2008alzheimer}. 
Pathological images are collected from 7 projects: TCGA-THYM, TCGA-THCA, TCGA-BRCA, TCGA-UCEC, TCGA-UVM, TCGA-OV, TCGA-MESO within the cancer genome atlas (TCGA) \footnote{\url{https://portal.gdc.cancer.gov/}}.

%
Detailed preprocessing steps for these upstream datasets are provided in Appendix \ref{sec:upstream pre}.

\noindent{\textbf{Downstream (DS) datasets}}:
Nine datasets covering five different modalities are used for downstream evaluation.
These include (1) Report: sentence classification (PudMed20k \cite{dernoncourt2017pubmed}); (2) X-ray: multi-class COVID-19 classification (ChestXR \cite{CovidGrandChallenge2021}) and COVID-19 infected region segmentation (QaTa-COV19-v2) \cite{degerli2022osegnet});
(3) CT: COVID-19 classification (RICORD \cite{tsai2021rsna}) and liver and liver tumor segmentation (LiTS \cite{bilic2023liver});
(4) MRI: vestibular schwannoma segmentation (VS \cite{shapey2021segmentation}) and left atrium segmentation (LA \cite{xiong2021global});
(5) Pathological imaging: colorectal cancer and healthy tissue classification (NCH \cite{kather2018100}) and gland segmentation (GlaS \cite{sirinukunwattana2017gland}).
More details about these downstream datasets are in Appendix \ref{sec:downstream pre}.

\begin{table*}[t]
\centering

    \setlength{\tabcolsep}{2pt}
  \caption{Generalization Performance of our MedCoSS, TFS, five single-modal pre-training models, and seven multi-modal pre-training on nine datasets. The best and second-best results in each column are highlighted in \textcolor[rgb]{ 1,  0,  0}{red} and \textcolor[rgb]{ 0,  .439,  .753}{blue}, respectively. *: Joint SSL with dimension-shared decoders. \dag: Joint SSL with modal-specific decoders.    }
\resizebox{1\textwidth}{!}{
    \begin{tabular}{c|ccc|ccc|cc|ccc|cc|cc|cc|ccc|cc}
    \toprule
    \multirow{2}{*}{Method} & \multicolumn{3}{c|}{PudMed20k (Report)} & \multicolumn{3}{c|}{ChestXR (X-ray)} & \multicolumn{2}{c|}{QaTa (X-ray)} & \multicolumn{3}{c|}{RICORD (CT)}  & \multicolumn{2}{c|}{LiTS (CT)} & \multicolumn{2}{c|}{VS (MRI)} & \multicolumn{2}{c|}{LA (MRI)} & \multicolumn{3}{c|}{NCH (Path.)} & \multicolumn{2}{c}{GlaS (Path.)} \\
\cmidrule{2-23}          
& ACC   & AUC & F1 & ACC & AUC   & F1  & Dice  & HD & ACC & AUC   & F1 & Dice  & HD  & Dice  & HD  & Dice  & HD  & ACC   & AUC   & F1    & Dice  & HD \\
    \hline
    TFS   & 82.41 & 94.82 & 76.45 & 82.36 & 94.32 & 81.55 & 73.94 & 38.38 &73.02 & 79.69 & 81.25  & 60.77 & 67.40 & 84.27 & 58.15 & 85.94 & 16.62 & 84.93 & 97.37 & 79.15 & 86.20 & 51.88 \\
    \hline
    \multicolumn{12}{l}{\textcolor{gray}{\textit{Single-Modal Pre-training}}} \\
    \hline
    Report & 83.79 & 95.33 & 77.97 & 79.73 & 92.67 & 79.04 & 72.31 & 40.29 & {74.21} & 81.42 & 82.36 & 59.71 & 63.41 & 43.17 & 151.83 & 83.92 & 17.45 & 86.31 & 97.65 & 81.07 & 85.73 & 56.12 \\
    X-ray & 82.40 & {94.80} & 76.55 & {\textcolor[rgb]{ 1,  0,  0}{95.55}} & \textcolor[rgb]{ 1,  0,  0}{99.21} & \textcolor[rgb]{ 1,  0,  0}{95.08} & \textcolor[rgb]{ 1,  0,  0}{80.27} & \textcolor[rgb]{ 1,  0,  0}{27.47} & {76.98} & 84.77 & 83.28 & 62.96 & 55.79 & 87.07 & 26.86 & 87.90 & 10.85 & 93.26 & 99.04 & 90.57 & 88.65 & 47.82 \\
    CT    & 82.59 & {94.91} & 76.84 & {85.37} & 95.63 & 84.69 & 74.40 & 36.10 & {\textcolor[rgb]{ 0,  .439,  .753}{80.56}} & 85.74 & \textcolor[rgb]{ 0,  .439,  .753}{86.09} & \textcolor[rgb]{ 0,  .439,  .753}{71.98} & \textcolor[rgb]{ 1,  0,  0}{33.79} & \textcolor[rgb]{ 0,  .439,  .753}{89.90} & \textcolor[rgb]{ 0,  .439,  .753}{10.91} & 87.21 & 16.01 & 88.25 & 98.31 & 83.79 & 87.42 & 50.81 \\
    MRI   & 82.82 & {94.93} & 76.96 & {84.14} & 95.14 & 83.47 & 74.08 & 36.22 & {79.37} & \textcolor[rgb]{ 0,  .439,  .753}{87.60} & 85.11 & 68.54 & 43.79 & 89.83 & 12.07 & 89.64 & \textcolor[rgb]{ 0,  .439,  .753}{10.24} & 88.29 & 98.40 & 83.72 & 87.33 & 52.13 \\
    Path. & 82.15 & {94.71} & 76.22 & {92.13} & 98.35 & 91.45 & 78.56 & 30.66 & {73.41} & 82.30 & 80.68 & 61.78 & 57.97 & 85.26 & 46.74 & 88.93 & 11.23 & \textcolor[rgb]{ 0,  .439,  .753}{95.62} & \textcolor[rgb]{ 1,  0,  0}{99.54} & \textcolor[rgb]{ 0,  .439,  .753}{94.01} & \textcolor[rgb]{ 1,  0,  0}{89.51} & \textcolor[rgb]{ 0,  .439,  .753}{46.79} \\
    \hline
    \multicolumn{12}{l}{\textcolor{gray}{\textit{Multi-Modal Pre-training}}} \\
    \hline
    Joint SSL* & \textcolor[rgb]{ 1,  0,  0}{84.10} & {\textcolor[rgb]{ 1,  0,  0}{95.52}} & \textcolor[rgb]{ 1,  0,  0}{78.30} & {92.24} & 98.33 & 91.59 & 78.61 & 30.29 & {77.78} & 85.89 & 84.13 & 70.59 & 38.24 & 89.73 & 15.33 & 89.41 & 13.84 & 94.63 & 99.27 & 92.93 & 87.86 & 51.20 \\
    Joint SSL\dag & \textcolor[rgb]{ 0,  .439,  .753}{83.94} & {95.33} & 78.13 & 91.54 & 98.02 & 90.85 & 78.29 & 30.27 & {78.97} & 86.65 & 84.90 & 69.60 & 44.17 & 89.41 & 15.83 & 88.84 & 14.94 & 94.48 & 99.30 & 92.69 & 87.83 & 52.08 \\
    EWC \cite{kirkpatrick2017overcoming}  & 83.39 & {95.17} & 77.66 & {90.48} & 97.77 & 89.72 & 77.44 & 31.67 & 75.79 & 82.18 & 82.59 & 64.59 & 53.66 & 88.15 & 23.87 & 88.00 & 12.55 & 94.50 & 99.37 & 92.61 & 88.16 & 50.81 \\
    ER  \cite{riemer2018learning}  & 83.75 & {\textcolor[rgb]{ 0,  .439,  .753}{95.43}} & \textcolor[rgb]{ 0,  .439,  .753}{78.15} & 85.32	&95.12	&84.43 &75.03 & 36.59 & {75.79} & 83.51 & 82.03 & 69.53 & 41.86 & 88.61 & 25.47 & 87.37 & 14.42 & 92.21  & 99.14   & 89.54   & 87.61 & 53.10 \\
    PackNet \cite{mallya2018packnet} & 82.88 & {95.01} & 77.06 & {86.14} & 95.98 & 85.28 & 75.69 & 35.98 & {71.43} & 83.75 & 79.63 & 61.89 & 62.12 & 83.92 & 54.07 & 88.87 & 12.55 & 92.42 & 99.13 & 90.00 & 86.96 & 52.81 \\
    CaSSLe \cite{fini2022self} &82.93	&95.12	&77.04	&91.66	&98.07	&90.99	&78.19	&31.23	&76.59	&82.85	&84.32  &67.04	&47.17	&87.94	&21.47	&\textcolor[rgb]{ 0,  .439,  .753}{89.66}	&10.67	&95.01	&99.44	&93.06	&89.12	&47.93 \\
    MedCoSS & 83.59 & {95.38} & 77.87 & {\textcolor[rgb]{ 0,  .439,  .753}{94.31}} & \textcolor[rgb]{ 0,  .439,  .753}{98.83} & \textcolor[rgb]{ 0,  .439,  .753}{93.77} & \textcolor[rgb]{ 0,  .439,  .753}{78.98} & \textcolor[rgb]{ 0,  .439,  .753}{29.43} & {\textcolor[rgb]{ 1,  0,  0}{83.33}} & \textcolor[rgb]{ 1,  0,  0}{88.74} & \textcolor[rgb]{ 1,  0,  0}{87.87} & \textcolor[rgb]{ 1,  0,  0}{72.01} & \textcolor[rgb]{ 0,  .439,  .753}{36.50} & \textcolor[rgb]{ 1,  0,  0}{90.12} & \textcolor[rgb]{ 1,  0,  0}{7.80} & \textcolor[rgb]{ 1,  0,  0}{90.46} & \textcolor[rgb]{ 1,  0,  0}{9.55} & \textcolor[rgb]{ 1,  0,  0}{95.76} & \textcolor[rgb]{ 0,  .439,  .753}{99.51} & \textcolor[rgb]{ 1,  0,  0}{94.01} & \textcolor[rgb]{ 0,  .439,  .753}{89.13} & \textcolor[rgb]{ 1,  0,  0}{46.69} \\

    \bottomrule
    \end{tabular}%
    }
    \label{tab:sota}
\end{table*}%

\begin{table*}[t]
  \centering
  \caption{Ablation studies of proposed feature distillation (FD), $k$-means sampling, and intra-modal mixup (IMM) strategies on five datasets. The baseline is the vanilla ER paradigm with a random sampling strategy. The best result in each column is highlighted with \textbf{bold}.
  }
  \resizebox{1\textwidth}{!}{
    \begin{tabular}{ccc|ccc|ccc|cc|cc|ccc|ccc|cc}
    \toprule
     \multirow{2}[4]{*}{FD} & \multirow{2}[4]{*}{$k$-means} & \multirow{2}[4]{*}{IMM} & \multicolumn{3}{c|}{PudMed20k (Report)} & \multicolumn{3}{c|}{ChestXR (X-ray)} & \multicolumn{2}{c|}{LiTS (CT)} & \multicolumn{2}{c|}{VS (MR)} & \multicolumn{3}{c|}{NCH (Path.)} & \multicolumn{3}{c}{Average (Cls)} & \multicolumn{2}{c}{Average (Seg)} \\
\cmidrule{4-21}          &       &       & ACC   & AUC   & F1    & ACC   & AUC   & F1    & Dice  & HD  & Dice  & HD  & ACC   & AUC   & F1    & ACC   & AUC   & F1    & Dice  & HD \\
    \midrule
    
          &       &       & \textbf{83.75}	&\textbf{95.43}	&\textbf{78.15}	&85.32	&95.12	&84.43	&69.53	&41.86	&88.61	&25.47	&92.21	&99.14	&89.54	&87.10	&96.57	&84.04	&79.07	&33.67 \\
    \checkmark &      &       & 83.53	& 95.25	& 77.68	& 92.16	& 98.27	& 91.43	& 71.20	& 37.56	& 89.89	& 11.70	& 95.17	& 99.42	& 93.16	& 90.29	& 97.65	& 87.43	& 80.54	& 24.63 \\
            &\checkmark       &       & 83.65	&95.36	&77.94	&84.72	&95.22	&83.85	&70.61	&38.31	&88.12	&31.46	&92.43	&99.15	&89.96	&86.93	&96.58	&83.92	&79.36	&34.88 \\

    \checkmark     & \checkmark     &       & 83.50	& 95.35	&77.77	&93.76	&98.76	&93.21	&71.56	&38.50	&89.95	&10.56	&95.47	&\textbf{99.57}	&93.88	&90.91	&97.89	&88.28	&80.76	&24.53 \\
    \checkmark     & \checkmark     & \checkmark     & 83.59 & 95.38 & 77.87 & \textbf{94.31} & \textbf{98.83} & \textbf{93.77} & \textbf{72.01} & \textbf{36.50} & \textbf{90.12} & \textbf{7.80} & \textbf{95.76} & 99.51 & \textbf{94.01} & \textbf{91.22} & \textbf{97.90} & \textbf{88.55} & \textbf{81.07} & \textbf{22.15} \\
    \bottomrule
    \end{tabular}%
    }
  \label{tab:ablation}%
\end{table*}%

\begin{table*}[t]
  \centering
  \caption{Performance of different pre-training orders on five datasets.}
  \resizebox{1\textwidth}{!}{
    \begin{tabular}{c|ccc|ccc|cc|cc|ccc|ccc|cc}
    \toprule
    \multirow{2}[0]{*}{Order} & \multicolumn{3}{c|}{PudMed20k (Report)} & \multicolumn{3}{c|}{ChestXR (X-ray)} & \multicolumn{2}{c|}{LiTS (CT)} & \multicolumn{2}{c|}{VS (MRI)} & \multicolumn{3}{c|}{NCH (Path.)} & \multicolumn{3}{c|}{Average (Cls)} & \multicolumn{2}{c}{Average (Seg)} \\
    \cmidrule{2-19}       
          & ACC   & AUC   & F1    & ACC   & AUC   & F1    & Dice  & HD95  & Dice  & HD95  & ACC   & AUC   & F1    & ACC   & AUC   & F1    & Dice  & HD95 \\
    \midrule
    Report, X-ray, CT, MRI, Path. (Default order) & 83.59 & 95.38 & 77.87 & 94.31 & 98.83 & 93.77 & 72.01 & 36.50 & 90.12 & 7.80  & 95.76 & 99.51 & 94.01 & 91.22 & 97.90 & 88.55 & 81.07 & 22.15 \\
    Path., MRI, Report, CT, X-ray & 83.63 & 95.27 & 77.79 & 95.70 & 99.25 & 95.24 & 71.43 & 36.58 & 89.29 & 11.93 & 93.75 & 99.30 & 91.19 & 91.03 & 97.94 & 88.07 & 80.36 & 24.26 \\
    \bottomrule
    \end{tabular}%
    }
  \label{tab:order}%
\end{table*}%

\subsection{Implementation details}
The default pre-training modality order in our pre-training setup is Report, X-ray, CT, MRI, and Pathological imaging.
We set the size of inputs as follows: the length of 1D vectors is set to 112, the size of 2D patches is set to $224\times224$, and the size of 3D patches is set to $16\times192\times192$. Augmentation techniques such as random crop, resize, and flip were applied to 2D images, while mirror augmentation was used for 3D images. Following MAE \cite{he2022masked}, the AdamW optimizer \cite{loshchilov2018fixing} was employed, setting the batch size to 512 and the maximum number of epochs per modality at 300. For each pre-training stage, a warm-up strategy was employed during the initial 40 epochs to gradually increase the learning rate from 0 to 1.5e-4, which was then decreased to 0 in the subsequent training according to the cosine schedule. The sampling number $K$ for each cluster, as detailed in Section \ref{method: kmeans}, was fixed at 5, indicating that we preserve 5\% of the previous data for rehearsal purposes.
During the fine-tuning phase, we continued with the AdamW optimizer but tailored the hyperparameters to suit the specific downstream tasks. Extensive details on these configurations were provided in Appendix \ref{sec:downstream pre}.
Performance in classification tasks was evaluated based on the area under the receiver operator curve (AUC), accuracy (ACC), and F1 score. For segmentation tasks, we employed the Dice similarity coefficient (Dice) and the 95\% Hausdorff distance (HD) as metrics.
To ensure the robustness of our conclusions, we averaged the results obtained using three different random seed values, including 0, 10, and 100.

\begin{figure*}[t]
  \centering
   \includegraphics[width=1.0\linewidth]{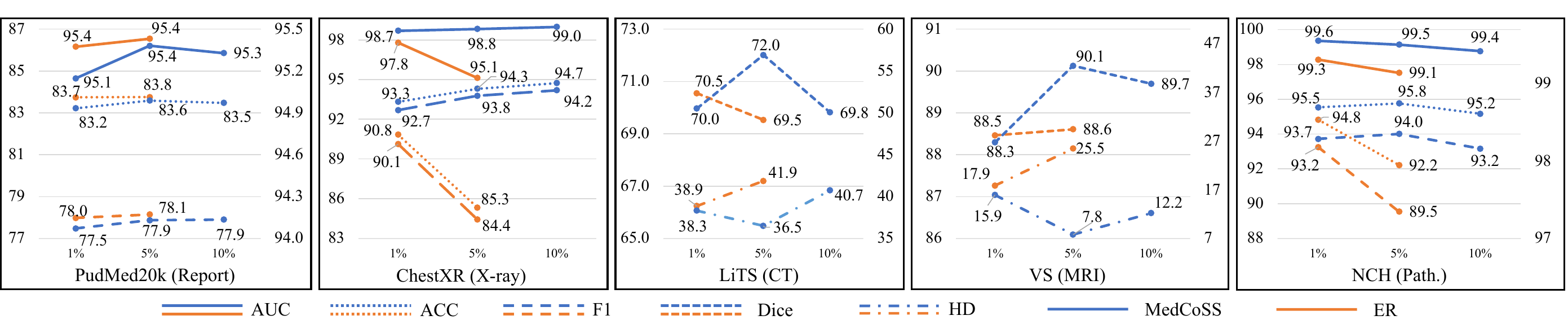}
   \caption{Comparison of our MedCoSS and ER with 1\%, 5\%, and 10\% buffer sampling ratio on five datasets. 
  }
   \label{fig:sampling ratio}
\end{figure*}

\begin{figure}[t]
  \centering
   \includegraphics[width=1.0\linewidth]{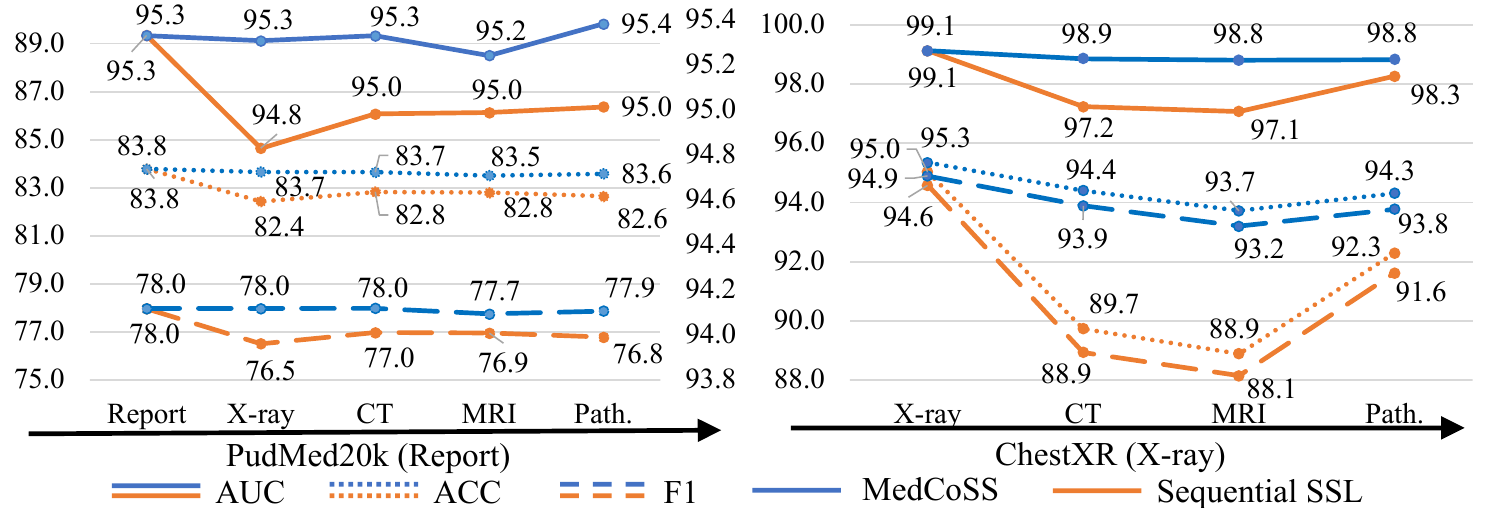}
   \caption{Comparison of our MedCoSS and sequential SSL for knowledge retention on the PudMed20k and ChestXR datasets. The sequential SSL means sequential pre-training without any knowledge retention strategies.}
   \label{fig:know reten}
\end{figure}

\subsection{Comparing to state-of-the-art SSL}
We comprehensively compared the proposed MedCoSS over other pre-training paradigms, including five single-modal pre-training paradigms (each trained on one of five modalities) and six multi-modal pre-training paradigms, across nine downstream datasets. The multi-modal paradigms are categorized into joint pre-training (including joint SSL with dimension-shared decoders and joint SSL with modal-specific decoders) and continual pre-training (including EWC \cite{kirkpatrick2017overcoming}, ER \cite{riemer2018learning}, PackNet \cite{mallya2018packnet}, and CaSSLe \cite{fini2022self}).
For all SSL paradigms, we employed masked modeling as the pretext task and ViT/B as the backbone with identical training epochs.
The performance of all those paradigms is displayed in Table \ref{tab:sota}. It first reveals that a well-pre-trained model achieves significant performance improvement over training from scratch (TFS).

\noindent{\textbf{Comparing to single-modal pre-training paradigms}}:
While single-modal pre-trained models demonstrate excellence on downstream datasets with identical modalities, their performance significantly drops on datasets with different modalities, particularly those of different dimensions, indicating limited generalization. For instance, the model per-trained on X-rays outperforms others on ChesXR and QaTa datasets but falls behind the model pre-trained on pathological images by 2.36\% in ACC and 3.44\% in F1 on the NCH dataset and the model pre-trained on CT scans by 3.58\% in ACC and 2.81\% in F1 on the RICORD dataset. 
In contrast, our model from MedCoSS exhibits more generalization performance, achieving the best or second-best results on eight out of nine datasets, with a marginal underperformance on the PudMed20k dataset compared to the model pre-trained on clinical reports.

\noindent{\textbf{Comparing to multi-modal pre-training paradigms}}:
We experimented with two joint SSL paradigms, \textit{i.e.}, joint SSL with dimension-shared decoders (maintaining three decoders for masking modeling) and joint SSL with modal-specific decoders (maintaining five decoders for masking modeling). 
Despite showing better generalization performance than single-modal pre-trained models, the models from both joint SSL paradigms often underperform compared to single-modal models when fine-tuning on identical modalities and our MedCoSS. 
Furthermore, they exhibit limited scalability in integrating new knowledge from stream-like data.
Among the other continual learning-based paradigms, including EWC, ER, PackNet, and CaSSLe, MedCoSS consistently achieves superior performance across eight datasets in all evaluated metrics.

\subsection{Ablation studies}
To validate the effectiveness of the proposed feature distillation (FD) strategy, $k$-means sampling strategy, and intra-modal mixup (IMM) strategy, ablation studies were conducted on five downstream datasets, each representing a different modality. 
In addition to the defined metrics, we calculated the average values of each metric for segmentation and classification tasks, respectively, to provide a holistic view of each strategy's effect. The results are displayed in Table \ref{tab:ablation}, where the first row reports the performance of the ER paradigm with a random sampling strategy \cite{riemer2018learning}. Based on these results, we have three conclusions: (1) adopting the FD strategy instead of the defined pretext task (\textit{i.e.}, masking modeling) markedly enhances knowledge retention and maintains the model's plasticity to new knowledge. As a result, the FD strategy improves average ACC, AUC, F1, Dice, and HD by 3.19\%, 1.08\%, 3.39\%, 1.47\%, and 9.04, respectively; (2) while the $k$-means strategy alone shows an unclearly positive impact, its combination with the FD strategy yields substantial gains in generalization across all datasets; (3) the integration of the IMM strategy furthers this improvement, achieving the best average results in both classification and segmentation tasks.

\subsection{Impact of buffer size on model performance}
In supervised continual learning, it is generally believed that a large buffer size equates to improved performance due to reduced forgetting \cite{buzzega2020dark, prabhu2020gdumb, tiwari2022gcr}.
For the multi-modal pre-training scenario, a larger buffer will bring the risk of intensified \textit{modal data collision} and computational cost, hindering current modality representation learning. To find an optimal balance, we experimented with ER and MedCoSS paradigms using sampling ratios of 1\%, 5\%, and 10\% across five datasets. 
The results are displayed in Fig. \ref{fig:sampling ratio}.
For the ER paradigm, a notable degradation in performance is observed when increasing the sampling ratio from 1\% to 5\%. On the contrary, the model from our MedCoSS demonstrates significant improvements across all datasets with the sampling ratio from 1\% to 5\% but shows performance degradation on PudMed20k, LiTS, VS, and NCH datasets except for the ChestXR dataset when the sampling ratio is raised to 10\%. 
Finally, we selected a 5\% sampling ratio for the rehearsal buffer in MedCoSS.

\subsection{Knowledge retention in sequential per-training}
While successfully mitigating the issue of the \textit{modal data collision}, the sequential pre-training paradigm inherently faces the risk of catastrophic forgetting. To address this, our MedCoSS paradigm incorporates three customized continual learning techniques. Fig. \ref{fig:know reten} illustrates the performance changes of both MedCoSS and a sequential SSL, where no knowledge retention strategies are employed, during the multi-stage pre-training process. This comparison distinctly shows that MedCoSS effectively preserves previous knowledge, exhibiting minor performance degradation over successive stages. In contrast, sequential SSL is observed to have significant sensitivity to catastrophic forgetting. For instance, after the CT pre-training stage, sequential SSL experiences a stark drop, dropping from 95.0\% to 89.7\% in terms of ACC. MedCoSS, on the other hand, maintains a more stable performance, with accuracy slightly decreasing from 95.3\% to 94.4\%.

\subsection{Pre-training order of modality}
We experimented with another set of modality pre-training orders: Pathological imaging, MRI, Report, CT, and X-ray. The results in Table \ref{tab:order} demonstrate that MedCoSS still achieves high generalization performance after the order change. 
Moreover, the modality placed last usually achieves better downstream performance than other positions. Due to the mitigated risk of knowledge forgetting in our MedCoSS, the performance difference between being last and other positions is insubstantial.

\section{Conclusion}
In this paper, we conduct an in-depth exploration of multi-modal SSL, specifically focusing on the practical but challenging unpaired data. We rethink the limitations in the prevalent joint SSL paradigm. Our experiments observed a so-called \textit{modal data collision}, which seriously hinders the representation learning across modalities. To alleviate this issue and ensure the scalability of the pre-trained model, we propose MedCoSS, a sequential pre-training paradigm that approaches multi-modal SSL through a multi-stage manner, assigning each stage to a specific modality. MedCoSS introduces a customized continual learning technique to alleviate inherent catastrophic forgetting within the sequential training scheme. We conduct experiments on a large-scale multi-modal unlabeled dataset, including the modality of Report, X-ray, CT, MRI, and Pathological imaging, and demonstrate the superiority of our MedCoSS on nine downstream datasets. 
In our future work, we plan to delve deeper into multi-modal medical pre-training within data streams, where each stage needs to handle specific multi-modal data.



\noindent{\textbf{Clinical impact:}}
Although MedCoSS may not currently achieve state-of-the-art performance across all tasks, its significant potential in advancing the multi-modal pre-trained medical universal model should be emphasized. The unique advantage of MedCoSS lies in its cost-effective scalability, which facilitates the seamless integration of new knowledge derived from real-world, stream-like data sources. This makes MedCoSS become a dynamic and adaptable paradigm, well-suited to keeping pace with the rapidly expanding variety of clinical multi-modal data.

{
    \small
    \bibliographystyle{ieeenat_fullname}
    \bibliography{main}
}

\clearpage
\setcounter{page}{1}
\maketitlesupplementary
\appendix

\begin{figure*}[t]
  \centering
   \includegraphics[width=0.90\linewidth]{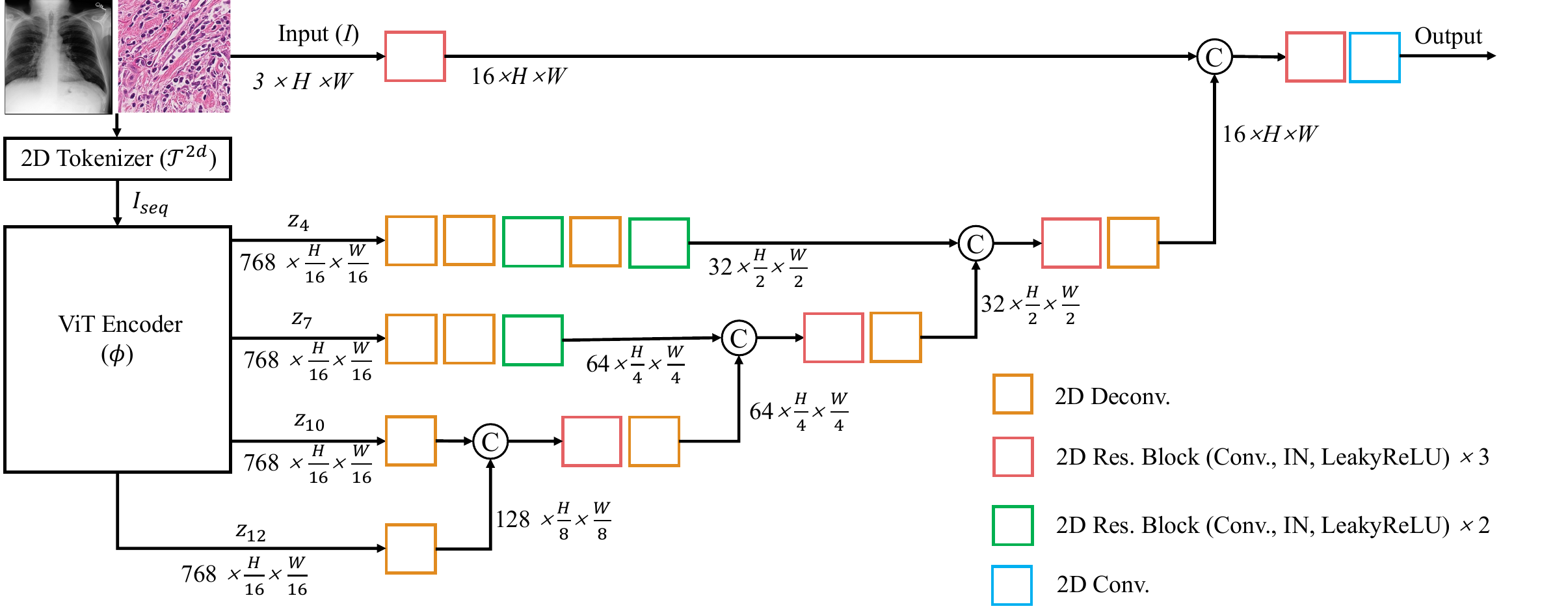}
   \caption{
   Illustration of model architecture for 2D segmentation tasks.
   }
   \label{fig: segnet}
\end{figure*}

\section{Architecture Details on Downstream Tasks} \label{sec:model details}
\subsection{Classification tasks}
We utilized a tokenizer chosen from the set $\{ \mathcal{T} ^{1d}, \mathcal{T}^{2d}, \mathcal{T}^{3d} \}$ based on the dimensionality of input data, the encoder $\phi$, and the newly appended Multi-Layer Perceptron (MLP) head to output the desired predictions. The architectures of these MLP heads are tailored to suit the specific requirements of individual datasets.

\noindent{\textbf{PudMed20k dataset and RICORD dataset}}:
A two-layer MLP head is adopted as the head. The first layer has 768 neurons followed by a layer normalization layer and the Gaussian error linear units (GELU). The second layer decreases the number of neurons to 5 and 2 for the PudMed20k and RICORD datasets, respectively.

\noindent{\textbf{ChestXR dataset and NCH dataset}}:
A one-layer MLP head is adopted as the head. The number of neurons is 3 and 9 on ChestXR and NCH datasets, respectively.

\subsection{Segmentation tasks}
We utilized a tokenizer chosen from the set $\{\mathcal{T}^{2d}, \mathcal{T}^{3d} \}$ based on the dimensionality of input data, the encoder $\phi$, and the randomly initialized decoder and segmentation head to produce prediction maps. Following \cite{hatamizadeh2022unetr,liu2023clip}, we devised 2D and 3D versions of decoders and segmentation heads to handle tasks in corresponding dimensions.

Considering a 2D image $I \in \mathbb{R}^{C \times H \times W}$ with resolution $(H, W)$ and $C$ input channels, it is divided into flattened, non-overlapping token sequences $I_{seq} \in \mathbb{R}^{N \times (P^{2}C)}$. Here, $(P, P)$ represents the resolution of each patch, and $N=(H\times W)/P^{2}$ denotes the total number of patches. These token sequences are then passed through the encoder, from which we extract four sequence features $\{z_{4}, z_{7}, z_{10}, z_{12}\}$, corresponding to the output of the 4-$th$, 7-$th$, 10-$th$, and 12-$th$ layers. These features $z_{i} \in \mathbb{R}^{\frac{H\times W}{P^{2}}\times 768}$ are subsequently reshaped to the shape of $768 \times \frac{H}{P} \times \frac{W}{P}$.
The bottleneck output of the encoder, \textit{i.e.}, $z_{12}$, along with the mid-stage output $z_{10}$, are processed by deconvolutional layers to upscale their resolutions. After concatenation, the obtained features are input into a residual convolution block to produce fused feature maps. Such a similar process is iteratively applied across all subsequent decoder layers up to the original input resolution where the last output is passed through a $1\times 1$ convolutional layer to generate pixel-wise segmentation predictions. More details were displayed in Fig. \ref{fig: segnet}.

For 3D image inputs, we employ a similar architecture, where the 2D tokenizer, convolutional layers, and normalization layers are replaced with their 3D counterparts, allowing for accepting 3D data.

\section{Dataset Details}
\subsection{Details of upstream datasets} \label{sec:upstream pre}
\noindent{\textbf{JPG version of the MIMIC-CXR 2.0.0 dataset}}:
This extensive public dataset \footnote{\url{https://physionet.org/content/mimic-cxr-jpg/2.0.0/}} comprises  377,110 JPG format chest radiographs and 227,827 clinical reports associated with these images. 
Following \cite{zhang2022contrastive, wang2022multi}, we excluded all lateral views from the dataset, as the downstream datasets only contain frontal-view chest images.
This resulted in a collection of 227,323 clinical reports and 356,309 chest radiograph images.

\noindent{\textbf{DeepLesion dataset}}:
This dataset includes 10,594 CT scans collected from 4,427 subjects \footnote{\url{https://nihcc.app.box.com/v/DeepLesion}}. Following \cite{ye2022desd}, each CT scan was resampled to a uniform spacing of $1.0 \times 1.0 \times 1.0$. We then applied cropping and resizing strategies to extract 125,070 sub-volumes. Specifically, we cropped each scan along the depth dimension using 24 window lengths and 12 strides, resizing the obtained sub-volumes to $16 \times 192 \times 192$.

\noindent{\textbf{ADNI dataset}}:
This dataset is a combination of ADNI-1, ADNI-2, and ADNI-GO datasets, utilizing a defined screening strategy \footnote{\url{adni.loni.usc.edu}}. The selection was based on diagnostic labels such as Alzheimer's disease, mild cognitive impairment, and cognitively normal, without specific consideration for sex, age, slice thickness, and manufacturer. Each MRI scan was cropped along the depth dimension using 16 window lengths and 8 strides, resizing the cropped volumes to $16 \times 192 \times 192$. In total, 59,205 sub-volumes were extracted.

\noindent{\textbf{TCGA dataset}}:
Comprising seven projects (TCGA-THYM, TCGA-THCA, TCGA-BRCA, TCGA-UCEC, TCGA-UVM, TCGA-OV, and TCGA-MESO) \footnote{\url{https://portal.gdc.cancer.gov/}}, this dataset includes a variety of pathological images. We processed these images by cropping them into non-overlapped $512\times512$ patches and resizing them to $224\times224$. For each patient with $n$ patches, we randomly selected $min(100, n)$ patches for training data.

\begin{table*}[h]
  \centering
  \caption{Implementation details of nine downstream tasks. CE: cross-entropy loss function.}
    \resizebox{1\textwidth}{!}{
    \begin{tabular}{ccccccccc}
    \toprule
    Dataset & Task Type & Modality & Loss Function  & Patch Size & Optimizer & Learning Rate & Batch Size & Iterations \\
    \midrule
    PudMed20k & Cls   & 1D Report & CE    & 112   & AdamW & 0.0002 & 64    & 14,065 \\
    ChestXR & Cls   & 2D X-ray & CE    & $224 \times 224$ & AdamW & 0.00005 & 32    & 35,840 \\
    QaTa  & Seg   & 2D X-ray & Dice + CE & $224 \times 224$ & AdamW & 0.0001 & 16    & 25,000 \\
    RICORD & Cls   & 3D CT & CE    & $64 \times 192 \times 192$  & AdamW & 0.00001 & 8     & 9,600 \\
    LiTS  & Seg   & 3D CT & Dice + CE & $64 \times 192 \times 192$  & AdamW & 0.0001 & 2     & 25,000 \\
    VS    & Seg   & 3D MR & Dice + CE & $64 \times 192 \times 192$  & AdamW & 0.0001 & 2     & 25,000 \\
    LA    & Seg   & 3D MR & Dice + CE & $64 \times 192 \times 192$  & AdamW & 0.00005 & 2     & 25,000 \\
    NCH   & Cls   & 2D Path. & CE    & $224 \times 224$ & AdamW & 0.0001 & 32    & 24,990 \\
    GlaS  & Seg   & 2D Path. & Dice + CE & $512 \times 512$ & AdamW & 0.0001 & 4     & 25,000 \\
    \bottomrule
    \end{tabular}%
    }
  \label{tab:setting}%
\end{table*}%

\begin{table*}[t]
  \centering
  \caption{Fine-tuning performance of five pre-trained models on four datasets, each representing different modalities. DeSD: 3D ResNet pre-trained on CT scans; Path\_DINO: ViT pre-trained on pathological images;  PCRLv2 (CheXpert/LUNA): 2D and 3D ResNets pre-trained on CheXPert (X-rays) and LUNA (CT scans), respectively; UniMiSS: Dimension-free pyramid U-like medical transformer (MiT) pre-trained on X-rays and CT scans. $\times$ means the model is incompatible with the dataset's data, resulting in an inability to be fine-tuned.
  }
    \resizebox{1\linewidth}{!}{
    \begin{tabular}{ccccccccccccc}
    \toprule
     \multirow{2}[0]{*}{Method}     & \multicolumn{3}{c}{PudMed20k (Report)} & \multicolumn{3}{c}{ChestXR (X-ray)} & \multicolumn{3}{c}{RICORD (CT)} & \multicolumn{3}{c}{NCH (Path.)} \\
     \cmidrule{2-13}  
          & ACC   & AUC   & F1    & ACC   & AUC   & F1    & ACC   & AUC   & F1    & ACC   & AUC   & F1 \\
    \midrule
     DeSD \cite{ye2022desd}  & $\times$    &  $\times$     & $\times$      & $\times$    & $\times$      & $\times$       & 78.57	& 83.46	 &  83.74   &  $\times$            & $\times$     & $\times$ \\
     PCRLv2 (LUNA) \cite{zhou2023unified} & $\times$      & $\times$      & $\times$      & $\times$      &  $\times$     & $\times$     & 81.35	&86.24	&86.62         & $\times$      & $\times$      & $\times$  \\
    Path\_DINO \cite{kang2023benchmarking} & $\times$     & $\times$     & $\times$     & 93.00 & 98.32 & 92.35 & $\times$     & $\times$     & $\times$     & 96.10 & 99.61 & 94.47 \\
    PCRLv2 (CheXpert) \cite{zhou2023unified} & $\times$     & $\times$     & $\times$     & 95.41 & 99.03 & 94.95 & $\times$     & $\times$     & $\times$     & 92.99 & 99.12 & 89.56 \\
    UniMiSS \cite{xie2022unimiss}  & $\times$     & $\times$     & $\times$     & 94.00 & 98.79 & 93.46 & 82.94 & 87.48 & 87.52 & 93.08 & 99.23 & 89.99 \\
    MedCoSS & 83.59 & 95.38 & 77.87 & 94.31 & 98.83 & 93.77 & 83.33 & 88.74 & 87.87 & 95.76 & 99.51 & 94.01 \\
    \bottomrule
    \end{tabular}%
    }
  \label{tab:pre-trained models}%
\end{table*}%

\begin{table}[t]
  \centering
  \caption{Performance of different backbones with different pre-trained models. CT: single-model SSL pre-training on CT data. The best result in each column is highlighted with \textbf{bold}.}
  \resizebox{1\columnwidth}{!}{
    \begin{tabular}{cc|cc|cc|cc}
    \toprule
    \multirow{2}[0]{*}{Method} &\multirow{2}[0]{*}{Backbone} & \multicolumn{2}{c|}{Liver} & \multicolumn{2}{c|}{Liver tumor} & \multicolumn{2}{c}{Average} \\
    \cmidrule{3-8}  
          & & Dice  & HD    & Dice  & HD    & Dice  & HD \\
    \midrule
    CT  & ViT/B  & 95.82 & 7.39  & 48.13 & 60.19 & 71.98 & 33.79 \\
    MedCoSS & ViT/B & 95.41 & 9.54  & 48.61 & 63.47 & 72.01 & 36.50 \\
    \midrule
    DeSD \cite{ye2022desd} & ResUNet & 96.81 & 4.24  & 65.27 & \textbf{26.84} & 81.04 & 15.54 \\
    CT + DeSD & ResUNet + ViT/B & 96.82 & 3.67  & 66.14 & 26.99 & 81.48 & \textbf{15.33} \\
    MedCoSS + DeSD & ResUNet + ViT/B & \textbf{96.90} & \textbf{3.27} & \textbf{66.79} & 27.84 & \textbf{81.84} & 15.56 \\
    \bottomrule
    \end{tabular}%
    }
  \label{tab:vit_resunet}%
\end{table}%

\subsection{Details of downstream datasets} \label{sec:downstream pre}
In Table \ref{tab:setting}, we provide the implementation details of nine downstream datasets. This includes information on the task type, modality, loss function, patch size, optimizer, learning rate, batch size, and maximum iterations. Additional information on each dataset is as follows:
(1) \textbf{PudMed20k dataset}. This dataset contains 20,000 abstracts from randomized controlled trials (RCTs), featuring a vocabulary of 68,000 words across 240,000 sentences. Each sentence is categorized into one of five labels: background, objective, method, result, or conclusion. We adopted the official data split, dividing the dataset into training, validation, and test sets, utilizing only the text data for category prediction.
(2) \textbf{ChestXR dataset}. This dataset focuses on detecting COVID-19, pneumonia, or normal in Chest X-ray images. It includes an official split of 14,958 training images and 3,432 test images, with 20\% of the training data randomly selected as the validation set.
(3) \textbf{QaTa-COV19-v2 (QaTa) dataset}. This dataset is utilized for COVID-19 infected region segmentation, with an official split of 7,145 training and 2,113 test images. A random 20\% of the training data serves as the validation set.
(4) \textbf{RICORD dataset}. This dataset contains 330 CT scans, which are divided into two categories: COVID-19 and normal.  All images were resized to $64\times 192 \times 192$. We followed the data split in \cite{xie2022unimiss}.
(5) \textbf{LiTS dataset}. This dataset includes 131 CT scans with annotations of liver and liver tumor segmentation, following the data split in \cite{ye2023uniseg}. The data were preprocessed by using nnUNet's preprocessing procedure \cite{isensee2021nnu}.
(6) \textbf{Vestibular-Schwannoma-SEG (VS) dataset}. This dataset contains 242 MRIs collected on patients with vestibular schwannoma. We followed the data split in \cite{ye2023uniseg}. The data were preprocessed by using nnUNet's preprocessing procedure \cite{isensee2021nnu}.
(7) \textbf{LA dataset}. This dataset provides 100 gadolinium-enhanced MRIs paired with left atrium ground truths. We followed the data split and preprocessing steps described in \cite{yu2019uncertainty}.
(8) \textbf{NCH dataset}. It consists of the NCT-CRC-HE-100K dataset (training set) and the CRC-VAL-HE-7K dataset (test set). We randomly sample 20\% of the training data as the validation set of each category.
(9) \textbf{GlaS dataset}. This dataset includes 165 pathological images from H\&E-stained colon tissue sections, labeled as malignant or benign. We adhere to the official data split and randomly sample 20\% of the training data for each category as the validation set.

\begin{figure}[t]
  \centering
   \includegraphics[width=0.80\linewidth]{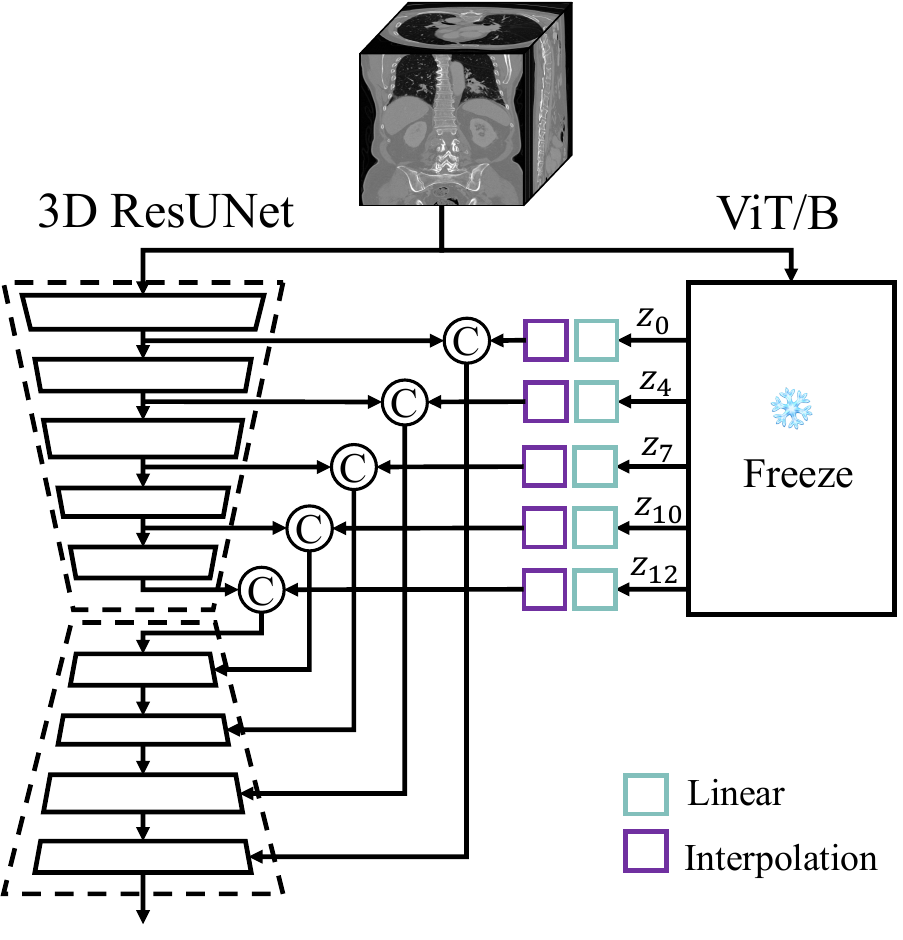}
   \caption{
   Illustration of the combination of ResUNet and ViT/B. The parameters of ResUNet and ViT/B are learnable and frozen, respectively.
   }
   \label{fig: vit_resunet}
\end{figure}

\begin{figure*}[t]
  \centering
   \includegraphics[width=1.0\linewidth]{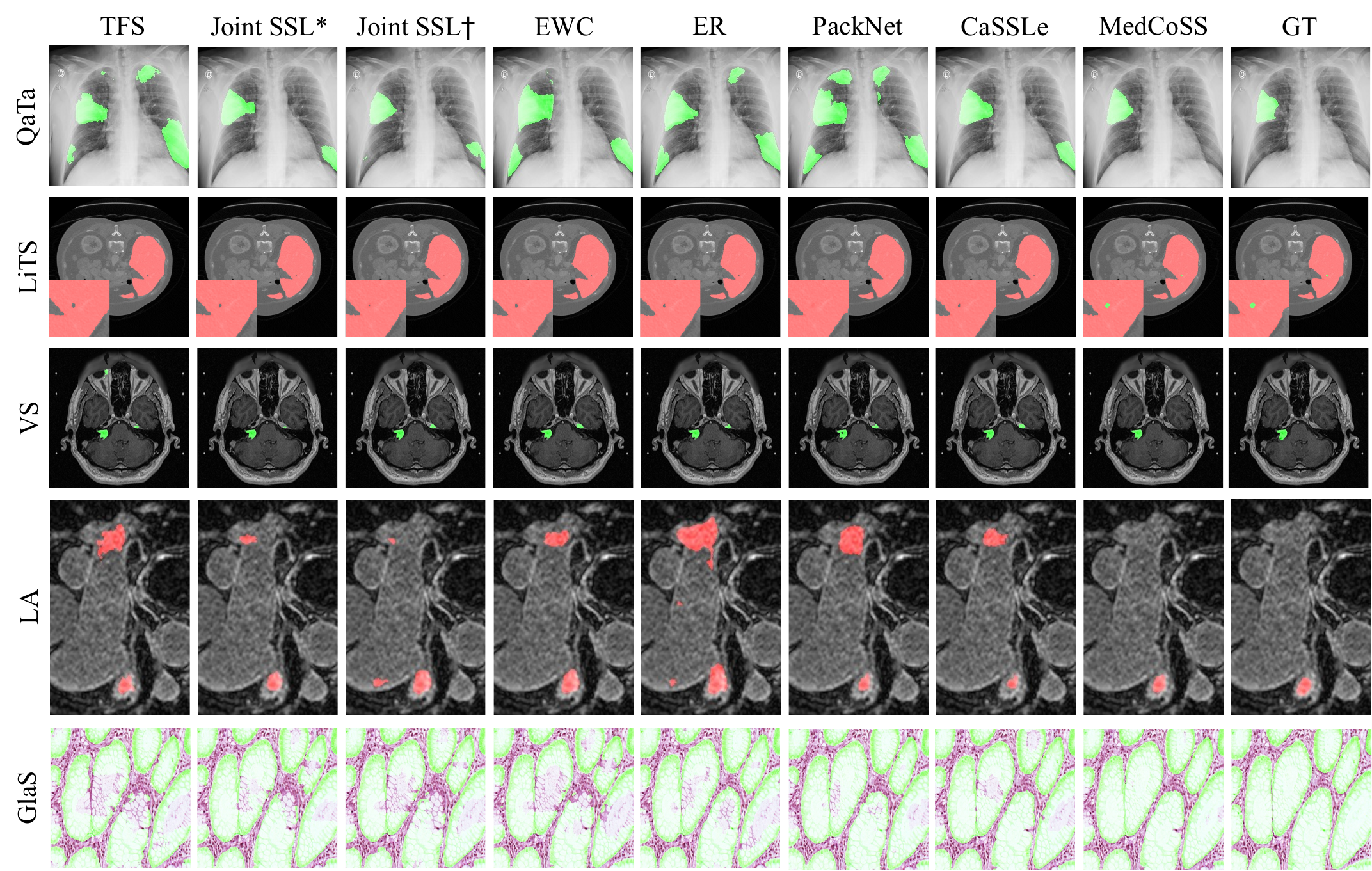}
   \caption{
   Visualization of segmentation results obtained by TFS, Joint SSL*, Joint SSL\dag, EWC, ER, PackNet, CaSSLe, and MedCoSS, and ground truths (GTs). The organs are colored red, and the tumors and malignant regions are colored green.
   }
   \label{fig: visualization}
\end{figure*}

\section{Comparing to Pre-trained Models}
We compared our MedCoSS over recent popular pre-trained models, including DeSD \cite{ye2022desd}, PCRLv2 \cite{zhou2023unified}, Path\_DINO \cite{kang2023benchmarking}, and UniMiSS \cite{xie2022unimiss}. All competing models were employed using their officially released weights and were fine-tuned for specific downstream tasks. The results shown in Table \ref{tab:pre-trained models} indicate that most models, except for the model from UniMiSS, are limited to handling data of a specific dimension.
These models demonstrate a significant decrease in performance when tasked with handling data from a different modality. 
For instance, Path\_DINO demonstrates superior performance across all metrics on the NCH dataset, yet it obtains the worst performance across all metrics on the ChestXR dataset. In contrast, our MedCoSS is versatile, accepting 1D, 2D, and 3D data, and consistently achieves robust generalization performance across all tasks.

\section{Improving 3D Segmentation}
While ViT-based models excel in uniformly handling multi-dimensional medical tasks and overcoming dimensional constraints, they tend to underperform in accurately segmenting small targets, such as tumors, which are prone to disappear or lost during passing through the tokenizer module. 
Inspired by nnSAM \cite{li2023nnsam}, we propose an integration of well-pre-trained ViT (obtained from MedCoSS or single-modal SSL) with CNN-based models (obtained from DeSD \cite{ye2022desd}).
This integration strategy, illustrated in Fig. \ref{fig: vit_resunet}, was tested on the LiTS dataset, containing liver and liver tumor segmentation annotations. The results in Table \ref{tab:vit_resunet} reveal that the pre-trained ResUNet obtained from DeSD achieves a significant performance gain compared to ViT-based models pre-trained by MedCoSS or single-modal SSL (CT). More importantly, the combination of ViT and ResUNet leads to further enhancement in performance.
This finding highlights the effectiveness of the integration, suggesting an alternative approach for leveraging the advantages of multi-modal pre-training to enhance the model's segmentation performance.

\section{Visualization of Segmentation Results}
For a qualitative comparison, we visualized the segmentation results derived from TFS, Joint SSL*, Joint SSL\dag, EWC, ER, PackNet, CaSSLe, and MedCoSS across five datasets, as shown in Fig. \ref{fig: visualization}. These visualizations reveal that the segmentation results from MedCoSS most closely align with the ground truths (GTs), effectively avoiding issues of over-segmentation and under-segmentation. For example, as shown in the second row of Fig. \ref{fig: visualization}, MedCoSS uniquely identifies a small liver tumor in the CT images, a task where the other paradigms fail.

\end{document}